\documentclass[sigconf,authorversion,screen,nonacm,timestamp]{acmart}

\usepackage{mathtools, amsmath, amsthm}

\usepackage{xfrac}
\usepackage{booktabs, dcolumn}
\usepackage{colortbl}
\usepackage{enumitem}
\usepackage{comment}

\usepackage{xcolor}

\definecolor[named]{BrickRed}{rgb}{0.71,0.2,0.11}
\definecolor[named]{NoteBlue}{rgb}{0.13,0.51,1.0}
\definecolor[named]{ForestGreen}{rgb}{0.0,0.7,0.2}

\newcount\Comments  
\definecolor{darkgreen}{rgb}{0,0.5,0}
\definecolor{darkred}{rgb}{0.7,0,0}
\definecolor{teal}{rgb}{0.3,0.8,0.8}
\definecolor{blue}{rgb}{0,0,1}
\definecolor{purple}{rgb}{0.5,0,1}
\newcommand{\kibitz}[2]{\ifnum\Comments=1\textcolor{#1}{#2}\fi}

\DeclareMathOperator{\FF}{FF}
\DeclareMathOperator{\Emb}{Emb}

\DeclareMathOperator{\softmax}{softmax}
\DeclareMathOperator{\bin}{bin}

\DeclareMathOperator{\tfidfc}{V_C}
\DeclareMathOperator{\tfidfw}{V_W}

\DeclareMathOperator{\CalL}{\mathcal{L}}    
\usepackage{float}

\newcommand{\entities}{\mathcal{E}}
\newcolumntype{$}{>{\global\let\currentrowstyle\relax}} 
\newcolumntype{^}{>{\currentrowstyle}}
\newcommand{\rowstyle}[1]{\gdef\currentrowstyle{#1}%
	#1\ignorespaces
}

\begin{document}

%

\title{Low Resource Recognition and Linking of Biomedical Concepts from a Large Ontology}

\author{Sunil Mohan}
\email{smohan@chanzuckerberg.com}
\affiliation{%
	\institution{Chan Zuckerberg Initiative}
	\city{Redwood City}
	\state{California}
	\country{USA}
	\postcode{94063}
}
\author{Rico Angell}
\email{rangell@cs.umass.edu}
\author{Nicholas Monath}
\email{nmonath@cs.umass.edu}
\author{Andrew McCallum}
\email{mccallum@cs.umass.edu}
\affiliation{%
	\institution{University of Massachusetts}
	\city{Amherst}
	\state{Massachusetts}
	\country{USA}
}


%





\begin{abstract}
Tools to explore scientific literature are essential for scientists, especially in biomedicine, where about a million new papers are published every year. Many such tools provide users the ability to search for specific entities (e.g. proteins, diseases) by tracking their mentions in papers. PubMed, the most well known database of biomedical papers, relies on human curators to add these annotations. This can take several weeks for new papers, and not all papers get tagged. 
Machine learning models have been developed to facilitate the semantic indexing of scientific papers. However their performance on the more comprehensive ontologies of biomedical concepts does not reach the levels of typical entity recognition problems studied in NLP. In large part this is due to their low resources, where the ontologies are large, there is a lack of descriptive text defining most entities, and labeled data can only cover a small portion of the ontology.
In this paper, we develop a new model that overcomes these challenges by (1) generalizing to  entities unseen at training time, and (2) incorporating linking predictions into the mention segmentation decisions. Our approach achieves new state-of-the-art results for the UMLS ontology in both traditional recognition/linking (+8 F1 pts) as well as semantic indexing-based evaluation (+10 F1 pts).
\end{abstract}

\maketitle

\section{Introduction}
\label{sec:intro}

The rate of publication in science research continues to grow. This is especially true in the biomedical and life sciences. The 2018 STM\footnote{International Association of Scientific, Technical, and Medical Publishers} report \cite{johnson2018stm} indicates that of the approximately 3 million scientific papers published in 2018 (as indexed by Elsevier's Scopus), the MEDLINE database of life sciences articles accounts for 850K\footnote{\url{https://www.nlm.nih.gov/bsd/medline_cit_counts_yr_pub.html}} ($\sim30$\%). Tools that facilitate efficient literature search and exploration have become vital for researchers \cite{landhuis2016scientific}. These tools (such as PubMed, Meta, and Google Scholar) provide users with the ability to set alerts and follow the mentions of particular scientific entities or topics in new publications. Additionally, the need for building efficient and effective biomedical literature exploration tools is of high public health importance, as seen with efforts such as CORD-19 to support the coronavirus pandemic \cite{wang2020cord}.

Semantic indexes that record the scientific entities (e.g. ``TAS2R38 protein, human'', ``Nipah Virus'', and the drug ``Atorvastatin'') mentioned in each paper are a core component of tools for literature search and exploration. These indexes account for ambiguity in the way entities are mentioned, and map them into a controlled ontology (e.g. the UMLS concepts C1175211, C0751673 and C0286651). This allows users to access information more easily without needing to query for a wide variety of aliases / spellings of an entity name.

To build such a semantic index, entity mentions must be recognized and linked in the text of scientific papers. These tasks are widely studied with many Wikipedia-based resources and a plethora of training data for newswire domains.
The domain of biomedical research papers, however, has a relative lack of training data resources. 
While datasets for some small ontologies have been widely studied, e.g. Chemicals and Diseases \citep{Li-etal:2016:BC5CDR}, 
the more useful comprehensive ontologies are large, lack descriptive text defining most entities, and labeled data can cover only a small portion of the ontology. At the same time, ontologies such as UMLS \cite{Bodenreider:2004} do contain useful information, supplementing entity names with aliases, acronyms, entity types and other information.

In this paper, we describe a new model for detection of mentions and linking them to concepts in an ontology of fine-grained types, that is designed to operate in a `{\em low resource setting}': (i) low coverage of the ontology in the training data, and (ii) with no descriptions for the concepts or their types. 

Inspired by the end-to-end linking model by \citet{Kolitsas-etal:2018}, we take a `bottom-up' approach, starting with considering all text spans as candidate mentions. These are then linked to concepts that best match the span in context. However, unlike previous work \cite{Kolitsas-etal:2018}, the final stage then predicts which of the linked spans are actual mentions in the document. Our approach is based on BERT \citep{Devlin-etal:2019:BERT-ACL}, taking advantage of its rich pre-training as well as leveraging its multi-level self-attention network to provide dynamic cross-attention between mentions and candidate concepts. We also leverage some basic information from the concept ontology. We thus avoid relying on specific trained concept (entity) or type embeddings, for which there are not enough resources in our problem setting.

We evaluate our approach on the largest public benchmark for biomedical entity recognition and linking
in research papers: MedMentions \cite{Mohan-Li:2019:AKBC}. 
This has been a particularly hard problem compared to other datasets for smaller biomedical ontologies; 
e.g. a simple BERT-based model for the easier NER task (recognizing mentions and their types but without linking) achieves SOTA performance with F1 scores above 0.85 and even as high as 0.935 for several other datasets \citep{Lee-etal:2019:BioBERT}. The best NER result on MedMentions ST21pv is F1 = 0.64 achieved by a recent much larger model \citep{Nejadgholi-etal:2019}. We attribute the low performance of SOTA models on MedMentions to its `low resource' nature, a setting our new model is designed to address.

We drastically improve the state-of-the-art for MedMentions in both the standard recognition and linking benchmarks, achieving a score 8 F1 points higher than the previous SOTA \cite{Loureiro-Jorge:2020:MedLinker}, while matching the best NER results with our linker. We also evaluate on the document-level linking metric, which is well aligned with the semantic indexing task and similarly provide a 10 F1 point improvement. Finally, we perform an extensive performance analysis of our method.

\section{End-to-End Linking Model}

\begin{figure*}
    \centering
    \includegraphics[width=\textwidth]{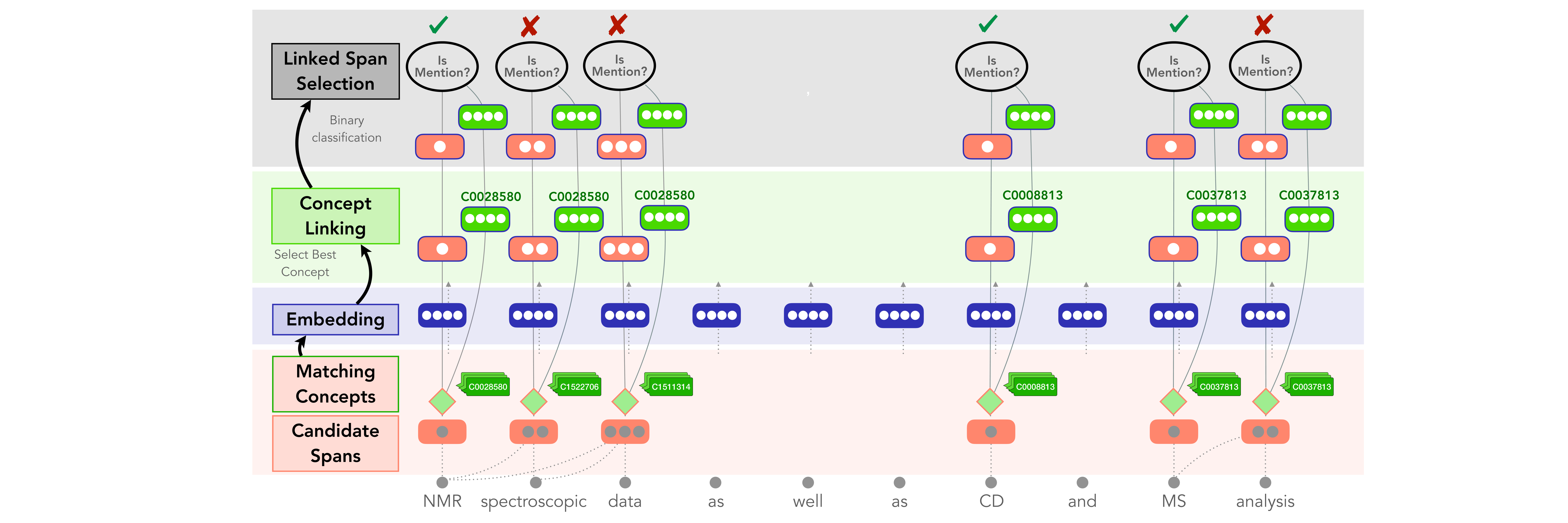}
    \caption{\textbf{Bottom-Up Architecture}: We start with all possible mention spans in the document, and lexically match them to the Entity KB. The Span Linker improves this match with contextualized semantics. The Span Selector takes the linking predictions and mention span representations, and determines which of the candidate spans and their linked entities should be labeled as concept mentions in the document. }
    \label{fig:arch_model}
\end{figure*}

Our proposed end-to-end recognition and linking model consists of three sequential stages: (i) \textbf{candidate generation} (section \S\ref{ssec:candidate_gen}): consider all text spans (of pretokenized text) as candidate mentions and use lexical similarity to generate matching entities from an alias table; (ii) \textbf{span linking} (\S \ref{ssec:concept_linking}): use contextualized semantic information to refine the lexical matches and select the best entities to link to each span; and finally, (iii) \textbf{span selection} (\S \ref{ssec:span_selection}): determine mention span boundaries using both the linking decisions and span representation.

Importantly, our model can generalize to entities that are \emph{unseen at training time}. It achieves this by using no entity-specific parameters; instead, entities are modeled by an encoding of their name spelling, type name, and other attributes. Additionally, the candidate generation and span selection components support generalization to new entities by considering all spans in the document.

\subsection{The Entity Recognition \& Linking Task}
\label{ssec:task_defn}

Given a document $D$ as a sequence of tokens, and a knowledge-base (KB) of entities $\entities$, we consider two tasks: (i) Entity Recognition and Linking requires predicting a set of mention-entity pairs $(m_{i}, e_i)$, where $m_{i}$ is a token span linked to the entity $e_i \in \entities$; (ii) Semantic Indexing requires predicting just the set of entities $\{e_i\}$ that are mentioned somewhere in the document.

UMLS is an ontology of biomedical concepts that are named fine-grained types. For convenience, we shall refer to these concepts as Entities, and the combined task of entity recognition and linking as \textit{concept recognition}.

\subsection{Candidate Mention Spans and Matches}
\label{ssec:candidate_gen}

Our model uses a high recall candidate phrase generator that generates a large number of overlapping candidate mention spans. Each span is then associated with a list of potentially matching entities.

We assume the knowledge base of entities $\entities$ is structured so that each entity is associated with an {\em entity type}, a preferred {\em primary name}, and a list of alternative names (aliases), each marked with a {\em name type} from an ordered list of name types (see \S \ref{sec:Preprocessing}). In general, the names are not unique, and $\entities$ is viewed as a list of entity alias entries $(a, E_a, T_e, t_a)$ where $a$ is a name, $E_a$ is an entity associated with that name, $T_e$ is a entity type, and $t_a$ is a name type.

The candidate generation step begins by considering all token spans of at most $K_S$ tokens contained within a sentence. We exclude spans that start or end with a stop word or punctuation token.

An enhanced version of the lexical matching scheme from \cite{Murty-etal:2018:HierarchicalLoss} is used to match candidate entities from $\entities$ to each candidate span. The tokenized candidate span $m$, and each name $a \in \entities$, are lemmatized and represented as TF-IDF vectors $\tfidfc(m), \tfidfc(a)$ of character $n$-grams and $\tfidfw(m), \tfidfw(a)$ of word $n$-grams. A lexical similarity score $S_M$ is computed between the span and each name, normalized to the range [-1, 1]:
\begin{align*}
%
s_M(m, a) & = \cos(\tfidfc(m) - \tfidfc(a)) + K_W \cdot \cos(\tfidfw(m) - \tfidfw(a)) \\
S_M(m, a) & = s_M(m, a) / (1 + K_W)
\end{align*}
The lexical matcher generates a list of matches 
$$LM(m) = [(S_M(m, a), a, E_a, T_e, t_a), \ldots]$$ 
which gets sorted on decreasing $(t_a, S_M(m, a))$ and only the first entry for each unique entity $E_a$ in the sorted list is retained. The output of the candidate generator is the top $K_M$ entries from this final list.

\subsection{Span Linking}
\label{ssec:concept_linking}

The span linker takes as input mention spans and their matches from the candidate generator, and is trained to predict the correct entity. The model uses BERT \cite{Devlin-etal:2019:BERT-ACL} to leverage contextualized semantics to match entities to mentions.

For a gold mention and matching candidate entity pair $(m, e),\ e \in LM(m)$ where $e=(s_e, a_e, E_e, T_e, t_e)$, the input sequence format is
\begin{equation*}
    \texttt{[CLS]}~C(m)~\texttt{[SEP]}~C_L(e)~\texttt{[SEP]}
\end{equation*}
where $C(m)$ is the mention centered in its textual context from the document and $C_L(e)$ is the textual form of the entity $e$.

The span linking model computes the logits
\begin{align*}
L(m, e) = \FF_L \bigl(  & B_L(m, e),\ \Emb(t_e), \\
	& \ s_e,\ \Emb(\bin_S(s_e)) \bigr)
\end{align*}
where $B_L$ is BERT's pooled output, $\Emb$ maps its input to trainable embeddings, $\bin_S(s_e)$ projects the lexical match score $s_e$ into bins, and $\FF_L$ is a feed-forward network. 
A trained vector is added to the embeddings of each mention token at BERT's input. The probability distribution across candidate matching entities for a mention is computed using $\softmax$, and the model is trained to optimize cross-entropy loss.

\subsection{Linked Span Selection}
\label{ssec:span_selection}

With refined linked entity predictions for each candidate span, the next step is to predict which spans are true mentions. 
The Linked Span Selection model takes as input all candidate mention spans from the Candidate Generator, and the top scoring $K_L \in \{1, \ldots, K_M\}$ entity matches with their predicted probabilities from the Span Linker. This model is similar to the Span Linker, except the second segment in the input sequence uses the representation $C_S(e)$ instead of $C_L(e)$.

For each input sample $(m,e,p)$, where $p = P_L(e | m)$ is the probability the entity $e$ is the right match for mention $m$ as computed by the Span Linker, the Span Selector computes the score:
\begin{align*}
S(m, e, p) = \FF_S \bigl(  & B_S(m, e),\ \Emb(t_e), \\
&  s_e,\ \Emb(\bin_S(s_e)), \\
&  p,\ \Emb(\bin_L(p)) \bigr)
\end{align*}
where $B_S$ is BERT's pooled output, and $\bin_L(p)$ projects the probability $p$ into predefined bins.
The model is trained to boost the score for correct linked mentions and suppress the score for in-correct inputs by optimizing a thresholded max-margin loss:
\begin{multline*}
\CalL_S(m, e, p)
= \begin{cases*}
W_{+} \times \max(0, M - s) & if $(m, e)$ is g.t. \\
\max(0, M + s) & o/w
\end{cases*}
\end{multline*}
where `g.t.' (`ground truth') marks positive samples, $s = S(m, e, p)$, $M$ the margin around threshold 0, and $W_{+}$ a weight applied to positive samples.

We evaluate performance of two inference modes for this model: the `Threshold' version selects all input samples with a score above a threshold (i.e. $S(m, e, p) > \tau$), and the `Greedy' mode adds a greedy selector that prefers higher scores of non-overlapping mentions that start early in the text. The Greedy mode works for the MedMentions dataset because it does not have any nested or overlapping mentions, otherwise the Threshold inference mode would be the better option.

\section{Model Specifics for Biomedical Text}

This section details the specifics of our approach specialized for biomedical concept recognition.

\subsection{MedMentions and the UMLS Ontology}

With our interest in automated semantic indexing of biomedical literature, we tested our model on MedMentions \cite{Mohan-Li:2019:AKBC}, the largest available dataset labeled with mentions of biomedical concepts. It consists of about 4,400 paper abstracts from PubMed, with mentions labeled with concepts from the UMLS 2017-AA ontology. 

Concepts in UMLS are named fine-grained types. Each concept is associated with a broader category called a {\em Semantic Type} (e.g. {\em C0029343:  Influenza A Virus, Avian} is associated with {\em T005: Virus}). UMLS provides each concept with a preferred {\em primary name}, and some known {\em synonyms} and {\em acronyms}.

There are about 3.2M concepts in UMLS 2017-AA Active.
As recommended in \cite{Mohan-Li:2019:AKBC} for semantic indexing,
we focus on the ST21pv subset of MedMentions, which restricts the concepts to a 2.3M large subset of the full ontology. Each concept in this subset is associated with one of 21 selected Semantic Types (the range of values for $T_a$ in \S~\ref{ssec:candidate_gen}). Despite the large number of papers, only $\sim1\%$ of the concepts are mentioned in the entire labeled corpus, and $\sim40\%$ of the concepts mentioned in the Validation and Test subsets are not mentioned in the Training subset.

\subsection{Pre-processing}
\label{sec:Preprocessing}

The MedMentions ST21pv corpus was processed as follows: (i) Abbreviations defined in the text of each paper were identified using AB3P \cite{Sohn-etal:2008:AB3P}. Each definition and abbreviation instance was then replaced with the expanded form (see Appendix~\ref{sec:appendix-A}). (ii) The text of each paper was tokenized and split into sentences using CoreNLP \cite{Manning-etal:2014:CoreNLP}. (iii) Overlapping mentions were resolved by preferring longer mentions that begin earlier. (iv) Finally, the corpus was saved in IOB2 tag format \citep{Kim-Veenstra:1999}.

The UMLS 2017-AA Active ontology was processed to build the knowledge base $\entities$ of concepts. All names were cleaned by removing supplementary text not likely to appear in a mention: (i) meta-information (e.g. ``{\em Formally}'', ``{\em Not Otherwise Specified}''); (ii) Disambiguating qualifiers (e.g. ``{\em Galanga $\langle$insect$\rangle$}'', distinguishes it from the plant ``{\em Galanga}'') were removed from the name, but saved for constructing a canonicalized name. Concepts were mapped to one of the 21 Semantic Types selected in ST21pv \citep{Mohan-Li:2019:AKBC} through the Semantic Type hierarchy, and unmapped concepts discarded. The resulting alias table contained 2,327,239 concepts and primary names, with 2,290,622 additional synonyms and 74,428 acronyms. Each name entry in $\entities$ was marked with one of the following ordered list of name types (i.e. $t_a$ in \S~\ref{ssec:candidate_gen}): Primary Name, Primary Name disambiguated, Acronym, Synonym.

\subsection{Textual Representations of Entities}

Entities are presented to the model in textual form as described in sections~\ref{ssec:concept_linking} and \ref{ssec:span_selection}. The span linker uses the canonicalized name $C_L(e) = {}$``$T_e\,,\ e_P\ e_C$'', where $T_e$ is the Semantic Type name (one of the 21 in ST21pv), $e_P$ is its Primary Name from UMLS, with any disambiguating context removed, and $e_C$ is its disambiguating context if present, enclosed in parentheses. For example for entity C4085630, the canonicalized name is ``\texttt{Eukaryote, Galanga (insect)}''. The span selector uses a similar representation: $C_S(e) = {}$``$T_e\,,\ a\ e_C$'', where $a$ is the alias name corresponding to the lexical match. See also Appendix \ref{app:linking_example} for more examples.

\section{Experiments}

We empirically compare our proposed model to state-of-the-art biomedical concept recognition (CR) approaches on the MedMentions \cite{Mohan-Li:2019:AKBC} dataset. We compare the performance across two metrics: (1) the standard mention-level metrics and, to better align with use of models as a semantic index, (2) document-level metrics. We also do an extensive analysis of our model, including its performance generalizing to unseen entities in the low-resource setting and its performance handling acronym mention spellings.

\subsection{Baseline Approaches}

We compare our proposed model to the following recent approaches:

\textbf{TaggerOne} \citep{Leaman-Zhiyong:2016} the baseline CR model for MedMentions reported in \citep{Mohan-Li:2019:AKBC}, does not use deep learning but also follows a bottom-up approach, using feature functions on lexical properties of the mention and its context, and does joint optimization of entity type recognition and entity linking. 

\textbf{MedLinker} \citep{Loureiro-Jorge:2020:MedLinker} is a recent NERL model that also uses a BERT-based NER stage followed by entity linking. They explicitly recognize the problem of low coverage of UMLS entities in training data by combining in the linker a classifier for entities seen during training, and an approximate dictionary matching stage for new entities. Their mention-level prediction metrics for MedMentions ST21pv exceed those of the baseline model in \citep{Mohan-Li:2019:AKBC} and were state-of-the-art before our model.

\subsection{Practical Details}

This section describes the various hyperparameters used in our model (summarized in \S~\ref{app:notation}). The character TF-IDF vector function $\tfidfc$ in the candidate generator represents the 200k most frequent character $n$-grams, $n \in \{2 \ldots 5\}$, $\tfidfw$ represents the 200k most frequent words in the names in $\entities$, and $K_W=0.5$. The output is limited to at most $K_M = 50$ matches for each span. The recall levels for various values of $K_M$ are shown in Table~\ref{tab:LexMatch-Recall}.

\begin{table}[t!]
\begin{center}
\begin{tabular}{$l^r^r}  
\toprule
$K_M$ & \textbf{Training} & \textbf{Validation} \\
\midrule
Recall@1   & 60.9\% & 61.8\% \\
Recall@50  & 85.8\% & 85.8\% \\
Recall@500 & 89.9\% & 90.1\% \\
\bottomrule
\end{tabular}
\end{center}
\caption{\label{tab:LexMatch-Recall}Lexical Match recall of correct Entity for gold mention spans, at different values of $K_M$.}
\end{table}

In the span linker model the bins used for $\bin_S$ were divided at $\{0, 0.2, 0.4, \ldots, 1.0\}$, and for $\bin_L$ in the  linked span selector at $\{0, 0.4, 0.5, \ldots, 0.9, 0.91, 0.92, \ldots, 1.0\}$. The embedding dimension used for these and the alias name type was 8. Both $\FF_L$ and $\FF_S$ were 3-layer feed-forward networks with hidden dimensions of 1024 and 256, {\em GeLU} \citep{Vaswani-etal:2017:NIPS} as the activation function and a dropout with probability 0.1 applied at their input. The length of the input sequence fed into BERT for both models was 128.

In the linked span selector we used $M = 1.0$, and in keeping with our focus on semantic indexing, we tuned $W_{+} \in \{1, 2, 5, 10, 20\}$ for the best document level F1 scores on the validation data.

Each input batch to the span linker consists of all candidates for one mention, with $K_M = 50$ as the batch size. The model was trained for up to 3 epochs with early stopping based on Recall evaluated on the validation data. For the Linked Span Selector, the input batch size is 64, and the model was trained for up to 10 epochs with early stopping based on F1 score for the validation data. Both models were trained using Adam with learning rates of 2$e$-5 (Span Linker) and 5$e$-6 (linked span selector), and a linear warmup for 10\% of the batches, followed by a linear decay.  

Both models use BioBERT-base-Cased (ver. 1.0) \citep{Lee-etal:2019:BioBERT} pre-trained on biomedical literature. BERT-based models use a lot of GPU memory, so for simplicity we tuned and trained each modeling stage separately.

We report the performance of our model for $K_L = 1$ and the score threshold $\tau = 0$.

\subsection{Evaluation}

We evaluate our approach along two metrics, following \citep{Mohan-Li:2019:AKBC}. The first is the standard mention-level performance using CoNLL chunking task metrics \citep{Sang-Buchholz:2000} measured against the tokenized and pre-processed documents.
To measure effectiveness for semantic indexing, we compute a document-level metric against the raw corpus; this evaluates the set of entities associated with each document without reference to the location of their mention in the text.

There are 203,282 mentions in the entire MedMentions ST21pv corpus of 4,392 documents. Tokenization results in a loss of 443 mentions, mostly due to resolving overlapping mentions. AB3P recognizes 7,376 abbreviation definitions in 2,941 documents. Abbreviation pre-processing (\S~\ref{sec:Preprocessing}) further reduces the total number of mentions in the tokenized corpus to 197,844, a total reduction of $\sim$2.7\%. However the essential structure and semantics of the text is not affected. The final mention predictions of the model correspond to the pre-processed versions of the documents.

\subsection{Model Performance}

\begin{table}[t!]
\begin{center}
\begin{tabular}{l^r^r^r}
\toprule
&	\bf Prec. & \bf Recall & \bf F1 \\
\midrule
\multicolumn{1}{r}{{\em Mention}:} \\
TaggerOne           & 0.471             & 0.436             & 0.453 \\
MedLinker           & 0.484             & 0.501             & 0.492 \\
*Our model (thresh.) & 0.630            & \textbf{0.520}    & \textbf{0.570} \\
*Our model (greedy)  & \textbf{0.650}   & 0.507             & \textbf{0.570} \\
\midrule
\multicolumn{1}{r}{{\em Document}:} \\
TaggerOne           & 0.536             & 0.561           & 0.548 \\
Our model (thresh.) & 0.682             & \textbf{0.633}  & \textbf{0.657} \\
Our model (greedy)  & \textbf{0.693}    & 0.621           & 0.655 \\
\bottomrule
\end{tabular}
\end{center}
\caption{\label{tab:ModelComparison}End-to-end metrics compared on Test data. (*)~This is a derived lower-bound by assuming null predictions for all mentions dropped during preprocessing.}
\end{table}

Table~\ref{tab:ModelComparison} compares the performance of our model against TaggerOne \citep{Mohan-Li:2019:AKBC} and MedLinker \citep{Loureiro-Jorge:2020:MedLinker}. For mention-level comparison, we derive metrics for our model against the raw corpus by assuming that all the mentions dropped during pre-processing receive a null entity prediction. This increases the number of false-negatives, lowering the Recall and F1 scores. Even with this conservative estimate, our model outperforms the others.

Document level metrics are only available for TaggerOne. Both `threshold' and `greedy' inference modes for our model perform better.


A detailed comparison of the two span selector inference modes can be seen in Table~\ref{tab:ModelMetrics}. Mention-level metrics are based on ground-truth from pre-processed documents, and document-level metrics use ground-truth from the unprocessed corpus. The `threshold' inference mode allows mentions to overlap. Since the `greedy' mode is a filter on the `threshold' selections, it lowers the recall while increasing the precision of the predictions, both at the mention and the document level. Finally, since Document-level metrics ignore mention span locations and evaluate only the set of entities predicted for each document, the Recall levels are higher than those for Mention-level predictions. Interestingly, the Precision scores are also higher.

\begin{table}[t!]
\begin{center}
\begin{tabular}{l^l^r^r^r}
\toprule
&	\bf Data & \bf Prec. & \bf Recall & \bf F1 \\
\midrule
\multicolumn{5}{c}{{\em Mention (on pre-processed documents)}:} \\
Threshold  & (validation) & 0.627            & \textbf{0.537}  & \textbf{0.579} \\
Greedy     & (validation) & \textbf{0.648}   & 0.523           & \textbf{0.579} \\
\cmidrule(l){2-5}
Threshold  & (test)       & 0.630            & \textbf{0.535}  & \textbf{0.579} \\
Greedy     & (test)       & \textbf{0.650}   & 0.522           & \textbf{0.579} \\
\midrule
\multicolumn{5}{c}{{\em Document (on raw ST21pv corpus)}:} \\
Threshold  & (validation) & 0.682            & \textbf{0.635}  & \textbf{0.657} \\
Greedy     & (validation) & \textbf{0.694}   & 0.624           & \textbf{0.657} \\
\cmidrule(l){2-5}
Threshold  & (test)       & 0.682            & \textbf{0.633}  & \textbf{0.657} \\
Greedy     & (test)       & \textbf{0.693}   & 0.621           & 0.655 \\
\bottomrule
\end{tabular}
\end{center}
\caption{\label{tab:ModelMetrics}End-to-end metrics on validation and test data.}
\end{table}

\subsection{Performance of Each Stage}

\begin{table}
\begin{center}
\begin{tabular}{l^r^r}
\toprule
					& \bf Training & \bf Validation \\
\midrule
Cand. Gen. ($K_M=50$)   & 0.853    & 0.853 \\
Linker ($K_L=1$)        & 0.952    & 0.851 \\
Span Sel. (thresh.)     & 0.826    & 0.740 \\
Span Sel. (greedy)      & 0.807    & 0.720 \\
\bottomrule
\end{tabular}
\end{center}
\caption{\label{tab:StagesRecall}Mention-level Recall for each stage, normalized to the ground-truth in their input.}
\end{table}

\begin{table}[t!]
\begin{center}
\begin{tabular}{l^r^r}
\toprule
 			& \bf Train & \bf Val. \\
\midrule
\#  Spans generated & 3.36M  & 1.13M \\ 
Recall of gold spans   & 0.993      & 0.991 \\
Recall of gold spans with & 0.853 & 0.853 \\
true entity in top 50 matches & & \\
\% matching gold spans  & 3.0\% & 3.0\% \\
\bottomrule
\end{tabular}
\end{center}
\caption{\label{tab:CandGen}Candidate Generator metrics.}
\end{table}

The candidate generator (Table~\ref{tab:CandGen}) generates over 99\% of the true mention (post-preprocessing) spans, and with $K_M = 50$, over 85\% of the true mentions have a matching candidate span with a matching entity. Since almost all text spans are considered as candidates, only 3\% of the candidate spans contain a true mention. Table~\ref{tab:LexMatch-Recall} shows recall levels at different values of $K_M$.

Table~\ref{tab:StagesRecall} gives the local Recall performance for the three model stages, where each stage is evaluated on what proportion of the ground-truth in its input is recalled in its output. In inference mode, the span linker is applied to all the candidate spans produced by the candidate generator (Table~\ref{tab:CandGen}). Its main effect is to increase the recall for the top-scoring entity prediction for each mention span from the candidate generator's 0.618 to 0.851 (validation data). Our end-to-end model cascades the three stages, so the final recall numbers are a product of the recalls for the three stages. 

The span selector is the weakest link in the sequence, with recall lower than the trained span linker model. For validation data, the recall levels at the input to the Span Selector model are 0.726 (mention) and 0.860 (document), which defines the upper-bound on span selector model's performance as reflected in the end-to-end metrics.

\subsection{Zero Shot Cases Evaluation}

Generalizing to `zero-shot' or {\em new} entities -- those unseen at training time -- is particularly important for using such a system as a semantic index in a production application that allows users to search for any entity in the knowledge base. One challenge in the MedMentions dataset is the large proportion of entities mentioned in the validation and test subsets that \emph{do not occur} in the training subset of the corpus. This is expected in a low-resource problem.

In MedMentions, 42.5\% of the set of unique entities mentioned in the test data are new. We evaluated the performance of our model on this dimension by comparing its predictions of mentions of new entities against new entity mentions in the test data, and correspondingly for {\em old} entities (those mentioned in training data). The metrics for the `greedy' inference mode are shown in Tables~\ref{tab:MetricsOldNew} and \ref{tab:CountsOldNew}. Despite the large proportion of new entities in the test data, the model recalls 43\% of their mentions (over 54\% of the entities at the document level). As expected, this is slightly lower than the recall level for old entities. Furthermore, the high precision of new entity predictions, and the low proportion of predictions that are for new entities, suggest that the model is reluctant to make predictions of new entities unless it has high certainty.

%
\begin{table}
\begin{center}
\begin{tabular}{l^l^r^r^r}
\toprule
	&            & \bf Prec. & \bf Recall & \bf F1 \\
\midrule
Unseen & (m)     & 0.858     & 0.430  & 0.573 \\
Seen   & (m)     & 0.659     & 0.548  & 0.598 \\
\cmidrule(l){2-5}
Unseen & (d)     & 0.959     & 0.513  & 0.668 \\
Seen   & (d)     & 0.708     & 0.648  & 0.677 \\
\bottomrule
\end{tabular}
\end{center}
\caption{\label{tab:MetricsOldNew}End-to-end metrics on Test data, Greedy inference, for entities previously {\em Seen} in Training v/s new {\em Unseen} entities: (m) Mention, (d) Document-level.}
\end{table}

%
\begin{table}
\begin{center}
\begin{tabular}{l^r^r^r}
\toprule
\multicolumn{1}{r}{{\em Entity type}:}      & \bf Unseen     & \bf Seen       & \bf Unseen\% \\
\midrule
True       & 8,715   & 30,323    & 22.3\% \\
Predicted  & 4,372   & 25,227    & 14.8\% \\
\bottomrule
\end{tabular}
\end{center}
\caption{\label{tab:CountsOldNew}Predicted and true mention counts in Test data of entities previously {\em Seen} in Training v/s {\em Unseen}.}
\end{table}

\begin{table}
\begin{center}
\begin{tabular}{l^r^r^r}
\toprule
\emph{Inference mode}  & \bf Prec.   & \bf Recall & \bf F1 \\
\midrule
Threshold       & 0.573     & 0.684  & 0.624 \\
Greedy          & 0.584     & 0.684  & 0.630 \\
\bottomrule
\end{tabular}
\end{center}
\caption{\label{tab:Acronyms}End-to-end mention-level metrics on Test data, for predicted matches to acronyms.}
\end{table}

\subsection{Evaluation -- Acronyms}

Some of the entity aliases in the ontology are acronyms, which are usually short and often ambiguous. The full set of entities for MedMentions ST21pv includes 67,593 unique (case-sensitive) acronyms, about half of which are 5 characters or shorter, and 4,596 acronyms map to multiple entities. Examples of ambiguous acronyms mentioned in the test data are ``{\em MS}'' which maps to 11 entities in the ontology (e.g. ``MS gene'', ``Mass Spectrometry'') and ``{\em ER}'' (7 entities, e.g. ``Endoplasmic Reticulum'', ``ESR1 gene'').

To evaluate the performance of our model on acronyms, we considered all predicted mentions whose predicted entity (from the Span Linker) was obtained by a lexical match to an alias name that is an acronym. We compared these predictions to all true mentions for the same spans. The results are in Table~\ref{tab:Acronyms}. As expected the Precision is lower than for all mentions (e.g. greedy precision 0.584 for acronyms is lower than the overall precision 0.650), but both Recall and F1 scores are significantly higher (e.g. greedy F1: $0.630 > 0.579$).

\subsection{Error Cases}

A detailed analysis of the model's false-positive mention predictions is shown in Table~\ref{tab:ErrorCases}. About 30\% of the cases have the correct span but not the correct entity. An interesting subset of this is where the span and semantic type are correct, but the entity is wrong (a fifth of the false-positives). Evaluating the predicted mention span and type, without the entity, measures {\em typed entity recognition} (aka NER). Both inference modes correspond to an NER F1 = 0.642. While this is computed on the pre-processed documents' mentions, even though our model is actually doing end-to-end linking, it compares favorably with the NER F1 of 0.64 reported in \citep{Nejadgholi-etal:2019}.

\begin{table}
\begin{center}
\begin{tabular}{l^r^r}
\toprule
\multicolumn{1}{r}{{\em Inference Mode}:}      & \bf Thresh.     & \bf Greedy  \\
\midrule
Correct Span, bad Entity & 28.7\%    & 31.3\% \\
Correct Span and Type     & 18.5\%    & 20.3\% \\
\cmidrule(l){1-3}
\multicolumn{3}{l}{\em Correct Entity, and True Span that:} \\
\ldots overlaps pred. Span      & 13.5\% & 12.5\% \\
\ldots contained in pred. Span  &  9.7\% &  9.5\% \\
\bottomrule
\end{tabular}
\end{center}
\caption{\label{tab:ErrorCases}Breakdown of false-positives in test data.}
\end{table}

The last two rows depict the proportion of false-positive predicted mentions whose span overlaps with or contains a true span with matching entity. These numbers explain why the document-level metrics are higher than the mention-level metrics.

\subsection{Robustness: Updating the Target Ontology}

Finally, we tested how well our model handled updates to the target ontology without re-training. For this we used UMLS 2020-AA, a more recent version of the UMLS ontology than the one (UMLS 2017-AA) used for training.  The ST21pv subset \citep{Mohan-Li:2019:AKBC} of UMLS 2020-AA adds 720k new concepts (an increase of 31\%) with a total of over 1M new names (+ 22\%) to the concept knowledge base (\S\ref{sec:Preprocessing}). The new ontology also drops some concepts from UMLS 2017-AA: 52 concepts mentioned in the test subset of MedMentions ST21pv were no longer part of UMLS 2020-AA.

We replaced the model's concept KB with this new knowledge base, and then generated concept predictions for documents in the test subset of MedMentions by running the model in inference mode as before. 
We then evaluated the model's document-level predictions using the Greedy inference mode, against the UMLS 2017-AA labels in MedMentions. The results are in table~\ref{tab:Comparison2020}. The new ontolgy adds many new concepts and concept names, while dropping a small number of the older concepts. Since the reference labels are in the original ontology, we expect a degradation in the metrics with the new ontology. The results show that this degradation is very small (-0.021 in F1).

\begin{table}
\begin{center}
\begin{tabular}{l^r^r^r}
\toprule
&	\bf Prec. & \bf Recall & \bf F1 \\
\midrule
UMLS 2017-AA (original)  & 0.693    & 0.621           & 0.655 \\
UMLS 2020-AA (new)       & 0.667    & 0.605           & 0.634 \\
\bottomrule
\end{tabular}
\end{center}
\caption{\label{tab:Comparison2020}Comparing the model's document-level Greedy predictions on Test data when using an updated ontology.}
\end{table}

We also wanted to evaluate how well the model recognized new concepts that were not present in the version of the ontology the model was trained with. For this we randomly selected 100 abstracts published in March from the LitCovid\footnote{\url{https://www.ncbi.nlm.nih.gov/research/coronavirus/}} dataset, to see how well our model recognized new concepts related to COVID-19. Since we did not have reference labeled data, we asked some expert biologists to do a quick evaluation of the model output. They were asked to rate each predicted concept on a 3-level scale: (i) Correct (is mentioned in the document), (ii) Related (is related to at least one mention in the document), or (iii) Incorrect. This is a much simpler criterion than that used in computing our precision-recall metrics; for example, the experts were not asked to evaluate how many concepts were missed, or whether the predicted concepts were the most specific concepts for that document.

As a reference, we also asked the biologists to perform the same task for predictions on 100 randomly selected documents from the test subset of MedMentions ST21pv, for both the original and new ontologies. The results are listed in table~\ref{tab:CurationResults}. They confirm that the degradation in performance when moving to a new ontology is quite small, even though the new ontology adds many new concepts and names. The model also performs quite well in recognizing concepts in documents from the LitCovid dataset, where 9.1\% of the predictions were for new concepts not present in UMLS 2017-AA.

\begin{table}
\begin{center}
\begin{tabular}{l^r^r^r}
\toprule
		                 &	\bf Test,     & \bf Test,     & \bf LitCovid, \\
UMLS version:            &	\bf 2017-AA   & \bf 2020-AA   & \bf 2020-AA \\
\midrule
\multicolumn{4}{l}{\em Nbr. of document-Concept predictions \ldots} \\
Total                   & 1,980           & 2,010         & 1,861 \\
New in UMLS 2020        &                 &    41         &   169 \\
\midrule
Correct Concepts         & 96.46\%        & 96.02\%       & 94.73\% \\
Related Concepts         &  2.02\%        &  2.29\%       &  3.76\% \\
Incorrect                &  1.52\%        &  1.69\%       &  1.50\% \\
\bottomrule
\end{tabular}
\end{center}
\caption{\label{tab:CurationResults}Expert Biologist's evaluation of predicted concepts, document-level using Greedy inference. Each tested corpus had 100 abstracts.}
\end{table}

\section{Related Work}
\label{app:related-work}

Concepts in UMLS are named fine-grained types, and each concept is also associated with a broader category called a Semantic Type in UMLS (e.g. {\em C0029343:  Influenza A Virus, Avian} associated with {\em T005: Virus}). This makes the task of Concept Recognition (CR) very similar to that of Named Entity Recognition and Linking (NERL).

The recognition of entities and concepts in natural language text is a widely studied task in NLP \cite[inter alia]{sang2003introduction,ratinov2009design,Lample-etal:2016} as is the linking of these mention spans to unambiguous knowledge-base entities \cite[inter alia]{cucerzan2007large,milne2008learning,ratinov2011local,ling2015design,raiman2018deeptype}. 

Even with the emergence of Deep Learning, most research in this field has focused on individual subproblems of NERL. Work on  Named Entity Recognition (NER), also known as (typed) Mention Detection, typically treats it as a problem of classifying individual tokens in the text on whether they belong to a mention of one of the targeted types \citep{Huang-etal:2015,Lample-etal:2016}. Similarly, work on Entity Linking (aka Entity Normalization / Disambiguation) takes as input a golden mention and a list of candidate matching entities with the goal of selecting the correct entity, usually taking advantage of entity descriptions available in Wikipedia \citep{FrancisLandau-etal:2016:BerkeleyCNN,Gupta-etal:2017:EMNLP}. The implication is that an end-to-end NERL system would follow this `{\em top-down}' approach, performing NER followed by EL.

Research on deep learning models for Biomedical CR has followed a similar pattern, mostly focusing on the NER subproblem and employing various combinations of CNN, LSTM and CRF  e.g. \citep{Li-etal:2015}, and language model based pre-training, e.g. \citep{Sachan-etal:2018}. The BERN end-to-end Biomedical NERL model \citep{Kim-etal:2019:BERN} applies a Bio-BERT based NER stage followed by rules and pre-existing type-specific linkers (Genes/Proteins, Diseases, Drugs and Mutations) or dictionary lookup (Species) for linking.

Taking a different perspective, the Large Scale Semantic Indexing Challenge tasks \citep{Nentidis-etal:2019:BioASQ} in the BioASQ\footnote{\url{http://bioasq.org}} workshops use a large dataset published by NCBI where documents are labeled with biomedical concepts from the MeSH ontology \citep{Lipscomb:2000:MeSH}, without identifying their mentions in the text. The ontology is much smaller than UMLS, with less than 30k concepts, and its coverage in the training data is above 95\%, compared to less than 1\% coverage in MedMentions, the dataset for our model. Leading approaches treat this as a multi-label document classification problem \citep{Peng-etal:2016:DeepMeSH,Xun-etal:2019:MeSHProbeNet}.


In a recent neural model for end-to-end NERL, \citet{Kolitsas-etal:2018} actually take a `{\em bottom-up}' approach, where all text spans are considered candidate mentions, and they jointly solve the problem of selecting the best mentions and linking them to correct entities. Their domain is named entities with Wikipedia pages, and they take advantage of the information available in Wikipedia, using internal hyperlinks to generate candidate entity matches to mentions, and training entity embeddings on their Wikipedia description pages to aid with the linking task. Our model also follows a bottom-up approach, however in our low-resource problem setting we do not have an external Wikipedia-like source of information with  hyperlinks or entity descriptions. We use lexical and string matching techniques similar to \citep{Murty-etal:2017} to generate candidate concept matches to a text span. 

The Zeshel model \citep{Logeswaran-etal:2019} presents an approach for zero-shot entity linking that allows for linking mentions to entities unseen at training time. However, this work also relies on entity description resources, and they do not do entity recognition, instead assuming that gold mentions are provided. Similar to Zeshel, our span linker and selector models use a trained vector to mark mention tokens. Instead of using entity descriptions, which are not available in our setting, we leverage the ontology.

TaggerOne \citep{Leaman-Zhiyong:2016}, the baseline CR model for MedMentions \citep{Mohan-Li:2019:AKBC}, does not use deep learning but also follows a bottom-up approach, using feature functions on lexical properties of the mention and its context, and does joint optimization of entity type recognition and entity linking. We use this as a baseline in our experiments.


The first model published on MedMentions \citep{Murty-etal:2018:HierarchicalLoss} addressed entity (concept) linking to gold mentions by taking advantage of the concept hierarchy in the UMLS ontology. They rely on learning concept embeddings as part of the model training, and use an unspecified subset of the MedMentions corpus. We do not consider this approach to training concept embeddings feasible even for the full ST21pv subset of MedMentions due to the low coverage of the ontology in the labeled data. Our model takes a more dynamic approach to matching entities to mentions, relying on a canonicalized name derived from the ontologycoupled with BERT-based cross-attention. 

In a more recent entity linking model \citep{Zhu-etal:2020:LATTE}, authors use a latent type modeling layer, with type prediction as an auxiliary task. Bi-directional attention flow is used to compute the interaction between the contextualized mention and concept names, combining this with latent type similarity to get the final score. On the full MedMentions corpus, authors report recall = 88.46 for the top-scoring candidate. However, the ground-truth entity is always included in the top 10 candidates. They use the same TF-IDF based candidate generator as \citep{Murty-etal:2018:HierarchicalLoss}; our candidate generator is an enhanced version of the same approach.

\citet{Nejadgholi-etal:2019} consider NER models for MedMentions, finding that a Bi-LSTM layer after BERT produces better results than a simple linear layer. Best results on MedMentions ST21pv (mention F1 = 0.64) are obtained by concatenating BERT-base and Bio-BERT \citep{Lee-etal:2019:BioBERT}. Authors also noted that $\sim 12$\% of the errors were the wrong type label on a correctly predicted text span. We show

MedLinker \citep{Loureiro-Jorge:2020:MedLinker} is a recent NERL model that also uses a BERT-based NER stage followed by entity linking. They explicitly recognize the problem of low coverage of UMLS concepts in training data, arguing that `traditional' approaches would not perform well on MedMentions. Accordingly, their linker stage combines a classifier for concepts seen during training, and an approximate dictionary match for new concepts. Their mention-level prediction metrics for MedMentions ST21pv were the SOTA, exceeding those of the baseline model in \citep{Mohan-Li:2019:AKBC}, as well as the traditional deep learning approach of ScispaCy \citep{Neumann-etal:2019:ScispaCy} (which achieves a mention-level F1 = 0.3424 on MedMentions ST21pv). We compare our results to MedLinker above, exceeding their performance by +8 F1 points.
\section{Conclusion}

In this paper, we present a new model for the recognition and linking of concepts in biomedical research papers. Our model is well suited for large low-resource ontologies which do not provide descriptions that semantically define the member entities, and most entities are not covered in the labeled training data. Our approach overcomes this challenge by starting with recall-biased candidate generation, uses dynamic entity encoding based on ontology features, and bases mention span selection on predicted entity links.

Our proposed approach achieves state-of-the-art results across two metrics of recognition and linking -- mention span level as well as semantic indexing evaluation using document level metrics  -- on the MedMentions dataset. We perform a detailed analysis of our method that studies the performance of the three components of our bottom-up model (candidate generation, linking, and span selection) as well as the performance on low resource entities and  ambiguous acronym mentions. We also demonstrate the robustness of our approach in its ability to handle ontology updates without new training data.

In future work, our model could be improved with better span selection, as that is the weakest link in the pipeline. Training a co-reference recognition component could improve recall for cases where the reference is indirect or too abbreviated to generate a good lexical match from the entity KB. A global cost model like that used in conditional random fields may also help improve span selection.

Better performance might also be possible by developing a smarter candidate span generator, and improving the recall of the lexical entity matcher. Finally, our model is sequential, and the errors from each stage cascade. Jointly optimizing the trained stages may result in improved performance.





%
%
\bibliographystyle{ACM-Reference-Format}

\bibliography{biblio}

\clearpage
\appendix

\section*{Low Resource Recognition and Linking of Biomedical Concepts from a Large Ontology: Supplementary Material}

\section{Abbreviation Processing}
\label{sec:appendix-A}


Here is an example to illustrate how abbreviations are processed. AB3P recognizes that a document defines ``{\em E. coli}'' as an abbreviation for ``\textit{Escherichia coli}'' in the text ``\ldots \textit{phosphorus and Escherichia coli (E. coli) in overland} \ldots''. After processing, this text containing the definition is replaced by just the expanded form: ``\ldots \textit{phosphorus and Escherichia coli in overland} \ldots''. If the abbreviated form was tagged as a mention, that mention also gets dropped. All other occurrences of the abbreviation ``\textit{E. coli}'' in that document are replaced with the expanded form ``\textit{Escherichia coli}'', and mention tags from the abbreviated form are copied to the inserted expansion.

\section{Hyper-parameters used in the model}
\label{app:notation}

\begin{table}
\small
\begin{center}
\begin{tabular}{$l^l}  
\toprule
\bf Parameter & \bf Description \\
\midrule
\multicolumn{2}{l}{\hspace{2em}{\em Candidate Generator}:} \\
$\tfidfc(\cdot)$   & Character $n$-gram TF-IDF vector \\
$\tfidfw(\cdot)$   & Word $n$-gram TF-IDF vector \\
$K_W$              & Weighting factor on $\tfidfw(\cdot)$ similarity \\
$K_M$              & Nbr. of top-ranked matches forwarded to the Linker \\
\cmidrule(l){1-2}
\multicolumn{2}{l}{\hspace{2em}{\em Span Linker}:} \\
$K_L$              & Nbr. of top linked matches forwarded to the Span Selector\\
$C_L(e)$           & Textual representation of entity (concept) $e$ \\
$\bin_S(s_e)$      & Projects the lexical match score $s_e$ into bins \\
\cmidrule(l){1-2}
\multicolumn{2}{l}{\hspace{2em}{\em Span Selector}:} \\
$C_S(e)$           & Textual representation of entity (concept) $e$ \\
$\bin_S(s_e)$      & Projects the lexical match score $s_e$ into bins \\
$\bin_L(p)$        & Projects the probability $p$ from the Linker into bins \\
$M$                & Margin in the max-margin loss function \\
$W_{+}$            &  Weight applied to positive samples in the loss function \\
$\tau$             &  Score threshold used during inference \\
\bottomrule
\end{tabular}
\end{center}
\caption{\label{tab:Notaion}Model hyper-parameters.}
\end{table}

Table \ref{tab:Notaion} lists the notation used in the paper for the main hyper-parameters for the model.

\section{Linking Ambiguous Mentions}
\label{app:linking_example}

\begin{figure}
    \centering
    \includegraphics[width=\columnwidth]{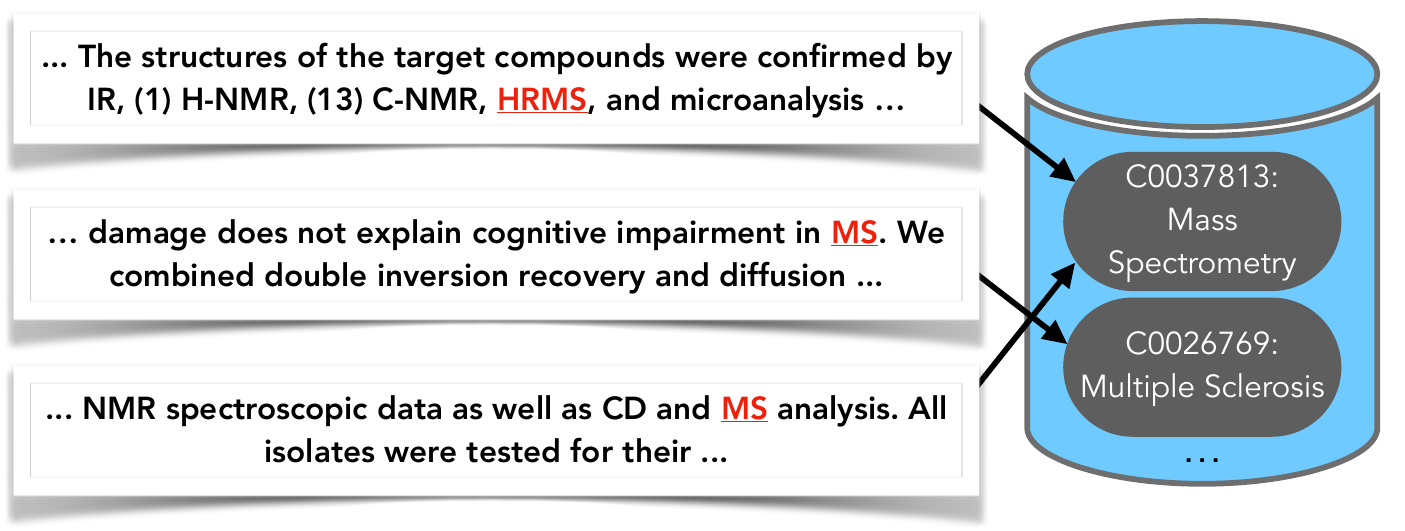}
    \caption{\textbf{Ambiguous Mention Linking}: An example showing 
    mentions with surface forms which could be attributed to multiple concepts in the ontology, or do not appear as an alias for any concept (``HRMS''). Our model is able to correctly link all of these mentions.}
    \label{fig:linking_example}
\end{figure}

Acronyms can often refer to multiple biomedical concepts. An example is ``MS'', which maps to 11 different entities in our targeted ontology. Figure \ref{fig:linking_example} shows some example mentions that are correctly linked by our model, including the acronym ``HRMS'' (an acronym for ``High Resolution Mass Spectrometry''), which actually does not occur as an alias for any of the entities in the ontology or entity KB. It gets linked correctly because of its lexical similarity to the known acronym ``MS'' for the entity ``Mass Spectrometry'', which also has a good semantic match to the context of that mention.

In these examples, the candidate generator finds lexical matches for ``HRMS'' and ``MS'' to the following concepts (among others):
\begin{itemize}
\item Type {\em T038: Biologic Function}, Concept {\em C0026769: Multiple Sclerosis}, matched to alias {\em MS}.
\item Type {\em T058: Health Care Activity}, Concept {\em C0037813: Mass Spectrometry}, matched to alias {\em MS}.
\end{itemize}
The textual representation $C_L(e)$ (see \S\ref{ssec:concept_linking}) of these concepts presented as input to the Span Linker are:
\begin{enumerate}
\item ``\texttt{Biologic Function , Multiple Sclerosis}''
\item ``\texttt{Health Care Activity , Mass Spectrometry}''
\end{enumerate}
The corresponding representation $C_S(e)$ (\S \ref{ssec:span_selection}) of these matches to the Span Selector are:
\begin{enumerate}
\item ``\texttt{Biologic Function , MS}''
\item ``\texttt{Health Care Activity , MS}''
\end{enumerate}

\end{document}